\pdfoutput=1
\documentclass[conference, a4paper]{IEEEtran}
\IEEEoverridecommandlockouts

\usepackage{cite}
\usepackage{amsmath,amssymb,amsfonts}
\usepackage{algorithmic}
\usepackage{graphicx}
\usepackage{textcomp}
\usepackage{xcolor}
\usepackage{subcaption}
\usepackage{amssymb}
\usepackage{amsmath}
\usepackage{tikz}
\usetikzlibrary{shapes,arrows}
\usetikzlibrary{arrows.meta}
\usetikzlibrary{angles,quotes}
\usetikzlibrary{3d}
\usepackage{amsmath,bm,times}
\usepackage{verbatim}
\usepackage{pgfplots}
\usepackage{multicol}
\usepackage{booktabs}
\usepackage{hyperref}
\usepackage{multirow}
\usepackage{subcaption} 

\usetikzlibrary{arrows.meta, positioning}
\usepackage{booktabs}
\usepackage{multirow}
\pgfdeclarelayer{background}
\pgfdeclarelayer{foreground}
\pgfsetlayers{background,main,foreground}

\tikzstyle{box}=[draw, fill=blue!20, text width=3em, 
            text centered, minimum height=2.0em, 
            font=\footnotesize]
\tikzstyle{ann} = [above, text width=3em]
\tikzstyle{box2} = [box, text width=4em, fill=red!20, 
    minimum height=5em, rounded corners]

\newcommand{\mx}[1]{\mathbf{\bm{#1}}} 
\newcommand{\vc}[1]{\mathbf{\bm{#1}}} 

\usetikzlibrary{positioning} 

\definecolor{pursuerblue}{HTML}{4c72b0}
\newcommand{\pursuertextcolor}[1]{\textcolor{pursuerblue}{#1}}
\definecolor{evaderorange}{HTML}{dd8452}
\newcommand{\evadertextcolor}[1]{\textcolor{evaderorange}{#1}}

\begin{document}

\title{Agile Interception of a Flying Target using Competitive Reinforcement Learning}

\author{\IEEEauthorblockN{Timothée Gavin}
\IEEEauthorblockA{\textit{IAS} \\
\textit{Thales LAS}\\
Rungis, France \\
timothee.gavin@thalesgroup.com}
\and
\IEEEauthorblockN{Simon Lacroix}
\IEEEauthorblockA{\textit{RIS} \\
\textit{LAAS CNRS}\\
Toulouse, France \\
simon.lacroix@laas.fr}
\and
\IEEEauthorblockN{Murat Bronz}
\IEEEauthorblockA{\textit{Dynamic Systems, OPTIM} \\
\textit{Fédération ENAC ISAE-SUPAERO ONERA, Université de Toulouse}\\
Toulouse, France \\
murat.bronz@enac.fr}
}

\maketitle

\begin{abstract}
This article presents a solution to intercept an agile drone by another agile drone carrying a catching net.
We formulate the interception as a Competitive Reinforcement Learning problem, where the interceptor and the target drone are controlled by separate policies trained with Proximal Policy Optimization (PPO).
We introduce a high-fidelity simulation environment that integrates a realistic quadrotor dynamics model and a low-level control architecture implemented in JAX, which allows for fast parallelized execution on GPUs.
We train the agents using low-level control, collective thrust and body rates, to achieve agile flights both for the interceptor and the target.
We compare the performance of the trained policies in terms of catch rate, time to catch, and crash rate, against common heuristic baselines and show that our solution outperforms these baselines for interception of agile targets.
Finally, we demonstrate the performance of the trained policies in a scaled real-world scenario using agile drones inside an indoor flight arena.
\end{abstract}


\section{Introduction}
The interception of agile aerial targets using autonomous drones is a challenging and increasingly relevant problem in robotics and security.
The increasing presence of unmanned aerial vehicles (UAVs) in unauthorized, restricted airspaces poses significant safety and security risks and has spurred interest in developing effective interception strategies \cite{Park2021}
In particular, scenarios such as airspace protection, infrastructure security, and event safety require the ability to capture or neutralize unauthorized drones with high precision and minimal collateral risk.
Deploying interceptor drones equipped with nets is a promising approach, but it demands advanced control capabilities to match or exceed the agility of evasive targets.

Traditional interception methods often rely on accurate models, preplanned strategies, or predictable target behaviour \cite{Yanushevsky2018}. 
However, modern quadrotor drones can perform highly dynamic manoeuvres, and will actively evade capture, rendering their trajectories unpredictable and challenging the effectiveness of classical methods \cite{Chung2011}.

\begin{figure}
    \centering
    \includegraphics[width=\linewidth]{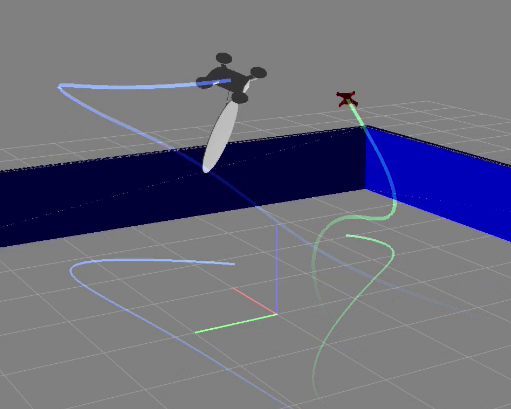}
    \caption{A competitive reinforcement learning approach to train both a pursuer and an evader drone for agile interception tasks. Both agents learn low-level control policies that enable them to perform dynamic maneuvers in a high-fidelity simulation environment.}
    \label{fig:cover_figure}
\end{figure}

Recent advances in deep reinforcement learning (RL) have demonstrated the potential to learn complex, high-dimensional control policies for drones directly from interaction with the environment. 
In particular in drone racing, RL-trained policies have achieved superhuman performance in highly dynamic and agile flight tasks \cite{Kaufmann2023champion}. 
However, the drone racing problem typically involve navigating static or slowly moving gates, while interception requires reacting to an adversarial agent that actively attempts to evade capture.

Competitive Multi-Agent RL (MARL) have shown outstanding results in adversarial settings, such as games \cite{Silver2017, OpenAI2019}. In this work, we formulate the agile interception problem as a competitive multi-agent RL task, where both the interceptor (pursuer) and the target (evader) are controlled by independent policies trained using Proximal Policy Optimization (PPO) in a co-evolution framework.
Our approach integrates a high-fidelity quadrotor dynamics model, enabling both agents to learn agile, physically realistic manoeuvres from low-level control inputs.
Through extensive simulation and real-world experiments, we show that RL training leads to robust and adaptive interception and evading strategies, outperforming heuristic control approaches.

The main contributions of this paper are:
\begin{itemize}
    \item A competitive MARL framework for agile drone interception, with both pursuer and evader learning from low-level control.
    \item Integration of a realistic quadrotor dynamics model to enable physically realistic and agile flight behaviors.
    \item Empirical evaluation demonstrating superior performance over standard baselines in simulation.
\end{itemize}

The remainder of the paper is organized as follows. Section \ref{sec:related_work} reviews related work on interception and agile flight.
Section \ref{sec:methodology} and \ref{sec:simulator} detail our agile flight simulation environment and our training methodology.
Section \ref{sec:results} presents experimental results and comparisons.
Section \ref{sec:conclusion} and \ref{sec:limitations} conclude and discuss future directions.


\section{Related-Work on the Agile Interception problem} \label{sec:related_work}

\subsection{Agile flight}\label{sec:agile_flight}

Agile flights in multi-rotors drones are typically characterized by the ability to perform large-angle manoeuvres, sustain high linear and angular accelerations, maintain precise control near dynamic limits and do so reliably in real-time, often in complex and cluttered environments.
Traditionally, achieving such agility relied on trajectory optimization coupled with controllers like Model Predictive Control (MPC), often requiring pre-planned paths and accurate system models \cite{mellinger2011minimum}.
However, these methods can be brittle when faced with unexpected disturbances of real-world flights.
Reinforcement Learning (RL) has emerged as a powerful alternative, enabling the learning of complex, non-linear control policies directly from interaction.
Research, such as work from \cite{Kaufmann2023champion} and \cite{Wang2024Dashing}, has demonstrated RL's capability to achieve highly dynamic and agile flights for quadrotors, pushing the boundaries of autonomous aerial manoeuvring beyond what traditional methods could easily achieve, particularly in tasks requiring aggressive, near-limits flight.

\subsection{Interception}\label{sec:interception}
Traditional interception methods uses heuristic or optimal control methods that often rely on accurate models, pre-planned strategies, or predictable target behaviour.\cite{Yanushevsky2018}. 
These approaches have been historically designed for the control of missiles and the interception of fixed-wing manned aircraft.
Optimal control methods requires accurate model of the pursuer and the evader to compute interception trajectories \cite{geisert2016trajectory}. Such model may not be available or may be too computationally expensive for real-time adaptation against unpredictable targets.
Heuristic guidance laws, such as Proportional Navigation (PN) or Pure Pursuit, offer computationally simpler alternatives and are widely used in missile guidance. However, these methods often assume relatively simple target manoeuvres and can struggle against highly agile or adversarial evaders.
Among recent works, \cite{pliska2024towards} have proposed heuristic methods for drone interception of agile manoeuvring targets, but these still assume a predictable target model.
More recently, learning-based solutions, particularly MARL have shown promise for developing complex control policies in adversarial settings.
MARL has been explored for pursuit-evasion games in various contexts, including simulated environments like Multi-Agent Particle Environments (MPE) \cite{Lowe2017} and initiatives like the DARPA AlphaDogfight Trials, demonstrating the potential to learn sophisticated tactics \cite{alphadogfighttrials2023}.
For quadrotors, \cite{Zhang2023} present a RL approach for quadrotor interception in an urban environment, and \cite{Chen2024} uses RL to give low-level commands for interception, however both these works consider only the pursuit side, assuming fixed evader behaviours.
The closest work to ours we found is \cite{Xiao2024} which uses RL to train both the pursuer and the evader in a co-evolution framework.
However, like most RL approaches \cite{Zhang2023}, they use high-level control inputs (e.g., velocity commands) and simplify the dynamics of the quadrotors, which limits the agility of the learned behaviours.
Overall, while RL has demonstrated potential in interception tasks, existing work either ignore the adversarial aspect of a learning evader, or often lacks the integration of highly dynamic capabilities in both the pursuer and the evader.

\subsection{Our approach}

Building upon the challenges highlighted in agile flight and interception, our approach directly addresses the need for highly dynamic capabilities in both the pursuer and the evader.
We recognize that interception is fundamentally a dynamic, adversarial interaction requiring controllers that can operate effectively near the physical limits of the hardware.
While RL has demonstrated remarkable success in achieving agile flight and has been applied to interception problems, to our knowledge, none have focused on training both an agile pursuer and an agile evader using low-level commands within a competitive RL framework.
Our work fills this gap by formulating the problem as a competitive multi-agent RL task where both agents learn physically plausible, agile manoeuvres through interaction in a realistic environment, to foster robust and adaptive strategies.


\section{Agile Flight Simulation Environment} \label{sec:simulator}

We use a high-fidelity simulator of the quadrotor dynamics. The simulator models air-drag, the low-level control architecture, the motor speeds, and the transmission delays.
This quadrotor model was taken from \cite{heeg2024learning} which include a low-level control architecture taking mass-normalized collective thrust and body rates as inputs.
We also implemented a high-level controller: an SE(3) controller \cite{lee2010geometric}, following the implementation from \cite{folk2023rotorpy}.
This controller, combined with the low-level quadrotor model, allow us to give alternate high-level commands to the quadrotors, such as position, velocity, or acceleration commands.
The control architecture is illustrated in Figure~\ref{fig:control_architecture}.
Our simulator also computes the collision between quadrotors, the elements of the arena and the net carried by the pursuers.
This fidelity facilitates the transfer of policies trained in simulation to the real world.

The simulation framework is entirely written using JAX \cite{Bradbury2018}.
This Python library allows the code to be just-in-time compiled and lowered to GPU during runtime, resulting in fast execution times of up to millions of steps per second by leveraging the parallelization capabilities of GPUs.

\tikzset{labelstyle/.style={font=\footnotesize, text width=2em} }

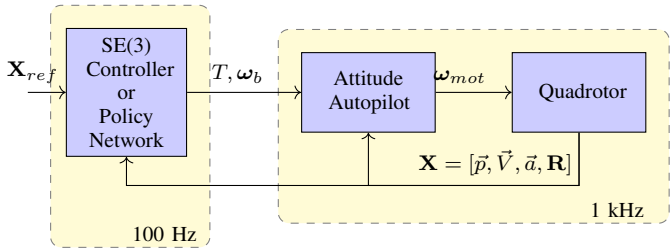
\begin{figure}[t]
  \centering
  \begin{tikzpicture}[
      node distance=1cm and 1.0cm, 
      auto
    ]

    \node (se3) [box, minimum width=4.5em, minimum height=3em] {SE(3) \\ Controller \\ or \\ Policy Network};
    \node (attitude) [box, right=of se3, xshift=0.5cm, minimum width=5em, minimum height=3em] {Attitude\\Autopilot};
    \node (vehicle) [box, right=of attitude, minimum width=5em, minimum height=3em] {Quadrotor};

    \node (X_ref) [left=of se3, xshift=0.5cm] {};

    \path [draw, ->] (X_ref.east) -- node [above, text width=3em, font=\footnotesize] {$\vc{X}_{ref}$} (se3.west |- X_ref) ;
    \path [draw, ->] (se3) -- ++(1cm,0) -- node [above, text width=3em, font=\footnotesize] {$T , \vc{\omega}_{b}$} (attitude.west |- se3) ;
    \path [draw, ->] (attitude) --  node [midway, above, text width=3em, font=\footnotesize] {$\vc{\omega}_{mot}$} (vehicle.west |- attitude) ;

    \node (feedback) [below=0.1cm of vehicle, xshift=-1.1cm, font=\footnotesize] {$\vc{X} = [\vec{p}, \vec{V}, \vec{a}, \mx{R}]$};

    \draw [->] (vehicle.south) -- ++(0,-0.7) -| (se3.south);
    \draw [->] (vehicle.south) -- ++(0,-0.7) -| (attitude.south);

    \node [below=0.8cm of se3, xshift=0.5cm, font=\footnotesize] {100 Hz};
    \node [below=0.8cm of vehicle, xshift=0.5cm, font=\footnotesize] {1 kHz};

    \begin{pgfonlayer}{background}
      \path (attitude.west |- attitude.north)+(-0.3,0.3) node (a) {};
      \path (vehicle.south -| vehicle.east)+(+0.3,-1.2) node (b) {};
      \path[fill=yellow!20,rounded corners, draw=black!50, dashed]
      (a) rectangle (b);
      \path (se3.north west)+(-0.2,0.3) node (a) {};
      \path (se3.south -| se3.east)+(+0.3,-1.2) node (b) {};
      \path[fill=yellow!20,rounded corners, draw=black!50, dashed]
      (a) rectangle (b);
    \end{pgfonlayer}

  \end{tikzpicture}
  \caption{Control architecture used for the quadrotor dynamics simulation.}
  \label{fig:control_architecture}
\end{figure}


\section{Interception of an Agile Target using Reinforcement Learning} \label{sec:methodology}

Drone neutralisation methods are classified as kinetic or non-kinetic \cite{Park2021}.
Non-kinetic approaches, such as jamming or spoofing, are ineffective against autonomous drones.
Kinetic methods, including projectiles or collisions, and electromagnetic weapons, risk causing uncontrolled crashes and debris.
Using a net, towed or projected by the pursuer, avoids these issues by safely capturing the target; here, we consider a net taut to the pursuer and released upon capture.

\subsection{Reinforcement Learning}

Reinforcement Learning is a type of machine learning for sequential decision-making.
In a \textit{rollout} phase, an \textit{agent} interacts with an uncertain \textit{environment} which provides it with a \textit{partial observations} of its \textit{state}, takes a series of \textit{actions} following a \textit{policy} and receives a scalar feedback in the form of \textit{rewards}.
These sequences of observe-act-reward, repeated over time, form the \textit{rollouts}.
The collected rollouts are then used to update the policy in a learning phase, which will then be employed in the rollout phase of the next training iteration.
The goal of the agent is to learn a policy that maximizes the expected cumulative reward over time.

Multi-Agent Reinforcement Learning (MARL) extends RL to scenarios with multiple agents interacting in a shared environment. 
MARL suffers from the curse of dimensionality and non-stationarity, as the environment dynamics change as other agents learn and adapt their policies.
Recent works in MARL adopted centralized training with decentralized execution (CTDE) \cite{Lowe2017}, where agents have access to global information during training but operate based on local observations during execution. In competitive settings, this alleviates non-stationarity by allowing the agents to access the state and actions of their opponents during training.

\begin{figure}[t]
  \centering
  \includegraphics[width=0.6\linewidth]{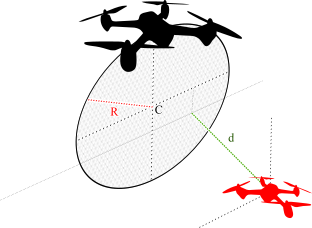}
  \caption{Schematic of the interception problem. Capture happens when the distance $d$ between the evader's centre comes within a capture distance of the pursuer's net.}
  \label{fig:caid_problem}
\end{figure}

\subsection{Pursuit-evasion problem}

We study a pursuit–evasion scenario with two quadrotors, a \emph{pursuer} and an \emph{evader}, operating in an obstacle-free rectangular arena of size \(L\times L\times H\).  
At the beginning of each episode, the agents’ initial positions are drawn uniformly at random inside the arena.  
The pursuer seeks to capture the evader as quickly as possible, whereas the evader tries to evade capture.  
Capture occurs when the evader’s centre comes within a \emph{capture distance} of the pursuer’s rigid, circular net of radius \(R\), which is mounted on the pursuer and aligned with its body frame, and represented in Figure~\ref{fig:caid_problem}.
Because a single pursuer cannot intercept a faster evader alone, we assume the pursuer's and the evader’s manoeuvring capabilities are identical.  
Target detection, state estimation, and trajectory prediction are not addressed in this work.

Both the evader and the pursuer are forbidden to exit the boundaries of the arena. 
In this setting the evader can quickly learn to fly close to the arena walls to stay safe, exploiting the pursuer’s fear to avoid boundary violations.
To discourage this behaviour and promote agile evasive flight in the central region, a narrow buffer zone is added adjacent to every wall; only the evader is penalized for entering this zone and thus, the evader is constrained in a smaller volume in the centre of the arena.

\subsubsection{Observation, actions, and rewards} 

At time step $t$, each agent's $i \in ({\rm pursuer, evader}) $ observation $\mathbf{o}_i$ is composed of the following elements: self-state observation $\mathbf{o}^{\rm self}_i$, observation of the opponent $\mathbf{o}^{\rm opp}_i$, and the observation of the arena bounds and the ground $\mathbf{o}^{\rm env}_i$.
The self-state observation is $\mathbf{o}^{\rm self}_i=\left[\mathbf{v}_i, \text{vec}\left(\mathbf{R}_i)\right)\right]$ containing the agent's linear velocity $\mathbf{v}_i$, and its rotation matrix $\mathbf{R}_i$, with $\text{vec}(\cdot)$ being the flattening function.
The observation of the opponent is ${o}^{\rm opp}_i=\left[\mathbf{p}_o-\mathbf{p}_i, \mathbf{v}_o-\mathbf{v}_i\right]$ containing the position and velocity of the opponent expressed relative to the agent in world coordinates.
The observation of the arena bounds and the ground is $\mathbf{o}^{\rm env}_i=\left[\text{norm}\left(\mathbf{p}_o-\mathbf{p}_i\right)\right]_{o\in \text{bounds $+$ ground}}$ and is composed of the Euclidean distances from the agent to each arena boundary and to the ground.
We normalize the observations before feeding them to the neural network. 
Relative positions are normalized by the maximum range of view $\mathbf{k_p}_i$, and velocities are normalized by a maximum velocity parameter for each agent $\mathbf{k_v}_i$.

The control policies are trained using Proximal Policy Optimization (PPO) \cite{Schulman2017}. This Actor-Critic method uses two neural networks for each agent: a policy network and a value network.

The policy network produces an action $\mathbf{a}_i$ for each agent, which is a vector of body rates $\mathbf{a}^\omega_i$ and a collective thrust $\mathbf{a}^{th}_i$.

The value networks are only used during training time and have access to privileged information about the opponent's state, which is not available to the policy network.
This alleviates the non-stationarity of the environment due to the simultaneous learning of both agents \cite{Lowe2017}.
The input of the value network of each agent is the concatenation of the position, velocity, and rotation matrix of each agent, as well as the action taken by the opponent at this time step. This input is normalized before being fed to the neural network.

The reward of the pursuer $r^{\text{P}}$ and the evader $r^{\text{E}}$ are given by:
\begin{align*}
  r^{\rm{P}} & = \phantom{-}r^{\rm{catch}} - r^{\rm{dist}} - r^{{\rm coll}} - r^{\rm{fail}} - r^{\rm{cmd}}, \\
  r^{\rm{E}}  & =             -r^{\rm{catch}} + r^{\rm{dist}} - r^{{\rm coll}} - r^{\rm{fail}} - r^{\rm{cmd}} - r^{\rm{bnd}}.
\end{align*}
in which $r^{\rm{catch}}$ rewards the pursuer for catching the evader,
$r^{\rm{dist}}$ penalizes the pursuer for being far from the evader,
$r^{\rm{fail}}$ penalizes any agent for crashing or going out of bounds, 
$r^{{\rm coll}}$ penalizes any agent for colliding with the body of their opponent, 
and $r^{\rm{cmd}}$ discourages dynamically infeasible commands.
Instead of terminating the episode upon collision between agents, we apply a soft continuous penalty $r^{{\rm coll}}$ to both agents, allowing for gradual learning of collision avoidance while maintaining focus on the primary tasks of pursuit and evasion.
We still terminate the episode if any agent crashes on the ground or goes out of bounds and apply a hard penalty $r^{\rm{fail}}$.
However, neither the evader nor the pursuer receive a reward when the opponent reaches a failure state to promote actual pursuit-evasion behaviours rather than forcing the opponent to crash.
Additionally, we add $r^{\rm{bnd}}$ to the evader's reward function, which penalizes it for approaching the arena bounds.

Specifically, the reward terms are:
\begin{align*}
  r^{\rm catch} &= \lambda_{\rm catch} \cdot \mathbf 1_{\rm catch},
  & r^{\rm dist} &= \lambda_{\rm dist}\cdot \bigl\lVert {\mathbf p}_e - {\mathbf c}^{\rm net}\bigr\rVert_2, \\
  r^{{\rm coll}} &= \lambda_{{\rm coll}}\, \mathbf 1_{\rm contact},
  & r^{\rm fail} &= \lambda_{\rm fail}\, \mathbf 1_{\rm fail}, \\
  r^{\rm cmd} &= \lambda_{\rm cmd}\, \lVert\mathbf a^\omega\rVert. 
  & r^{\rm bnd} &= \phi_{\rm{bnd}}(d^{\rm bnd}),
\end{align*} in which the indicator functions return $1$ when their condition is met : 
$\mathbf 1_{\rm catch}$ when catching the evader,
$\mathbf 1_{\rm contact}$ for inter-agent contact, 
$\mathbf 1_{\rm fail}$ for reaching a failure state because of a ground crash or leaving the arena bounds. 
$c^{\rm net}$ is the pursuer's catching net-centre position,
and $\mathbf{a}^\omega$ are the commanded body rates.
$\phi_{\rm bnd}$ is a function that penalizes the evader for approaching the arena bounds, 

triggering under a set threshold and growing exponentially the shorter the distance to the arena bounds $d^{\rm bnd}$.
$\lambda_{\rm catch}$, $\lambda_{\rm dist}$, $\lambda_{{\rm coll}}$, $\lambda_{\rm term}$, $\lambda_{\rm cmd}$ are positive hyperparameters that balance the different reward terms and have been tuned to obtain the desired behaviour and listed in Table~\ref{tab:reward_coefficients}.

\begin{table}[h]
  \centering
  \caption{Reward coefficients.}
  \label{tab:reward_coefficients}
  \setlength{\tabcolsep}{6pt}
  \renewcommand{\arraystretch}{1.1}
  \begin{tabular}{@{}lclc@{}}
    \toprule
    \textbf{Coefficient} & \textbf{Value} & \textbf{Coefficient} & \textbf{Value} \\ 
    \midrule
    $\lambda_{\rm catch}$ & 10.0 & $\lambda_{\rm coll}$ & 0.1 \\
    $\lambda_{\rm dist}$ & 0.001 & $\lambda_{\rm fail}$ & 30.0 \\
    $\lambda_{\rm cmd}$ & 2e-04 & $\lambda_{\rm bnd}$ & 1.0 \\
    \bottomrule
  \end{tabular}
\end{table}

\begin{table}[b]
  \centering
  \caption{Training hyperparameters.}
  \label{tab:hyperparameters}
  \setlength{\tabcolsep}{6pt}
  \renewcommand{\arraystretch}{1.1}
  \begin{tabular}{@{}lc@{}}
    \toprule
    \textbf{Hyperparameter} & \textbf{Value} \\ 
    \midrule
    Number of parallel environments (per agent) & 1024 \\
    Rollout length & 128 \\
    Learning rate & $5\times10^{-4}$ \\ 
    Discount factor & 0.99 \\ 
    Number of PPO epochs per training data batch & 15 \\
    Number of minibatches per PPO epoch & 1 \\
    Discount factor & 0.99 \\
    Lambda value for GAE computation & 0.95 \\
    Clipping value for PPO updates & 0.2 \\
    Entropy & 0.01 \\
    Critic weight in loss function & 0.5 \\
    Maximum norm of the gradients for a weight update & 0.5 \\
    Decay learning rates & False \\
    Total number of training steps & $4\times10^9$ \\
    \bottomrule
  \end{tabular}
\end{table}
\subsection{Training details}\label{sec:training}

Rollouts are generated in parallel across $1024$ environments.  
Episodes start from uniformly sampled initial positions in the $L\times L\times H$ arena; no domain–randomisation of the platform dynamics is applied.
Episodes last up to $T=10$\,s (1000 time steps) unless terminated earlier due to capture, crash, or arena exit.

Each policy network is a two‐layer multilayer perceptron with $256$ ReLU units per hidden layer.  
The output layer produces the mean and standard‐deviation of a multivariate Gaussian, followed by a \texttt{tanh} squashing to obtain bounded continuous actions.  
The value networks mirrors this architecture but ends with a linear output.

The entire pipeline is written in Python using JAX \cite{Bradbury2018}, enabling just‐in‐time compilation and parallelized execution.
Running on a single machine equipped with an NVIDIA RTX~4090 (24\,GB VRAM), an AMD Ryzen 9 7950X3D (16~cores, 4.2\,GHz) and 128\,GB RAM, the system collects and processes approximately \(3.5\times10^{5}\) environment steps per second.
We train for a total of \(2\times10^{9}\) environment steps, corresponding to roughly 1h35 of wall-clock training time.
Training hyperparameters are listed in Table~\ref{tab:hyperparameters}.


\section{Experimental Results}
\label{sec:results}

\subsection{Training Results}

\begin{figure}[t] 
  \centering
  \begin{subfigure}[t]{0.48\textwidth}
    \includegraphics[width=\textwidth,clip,trim=0 100 0 200]{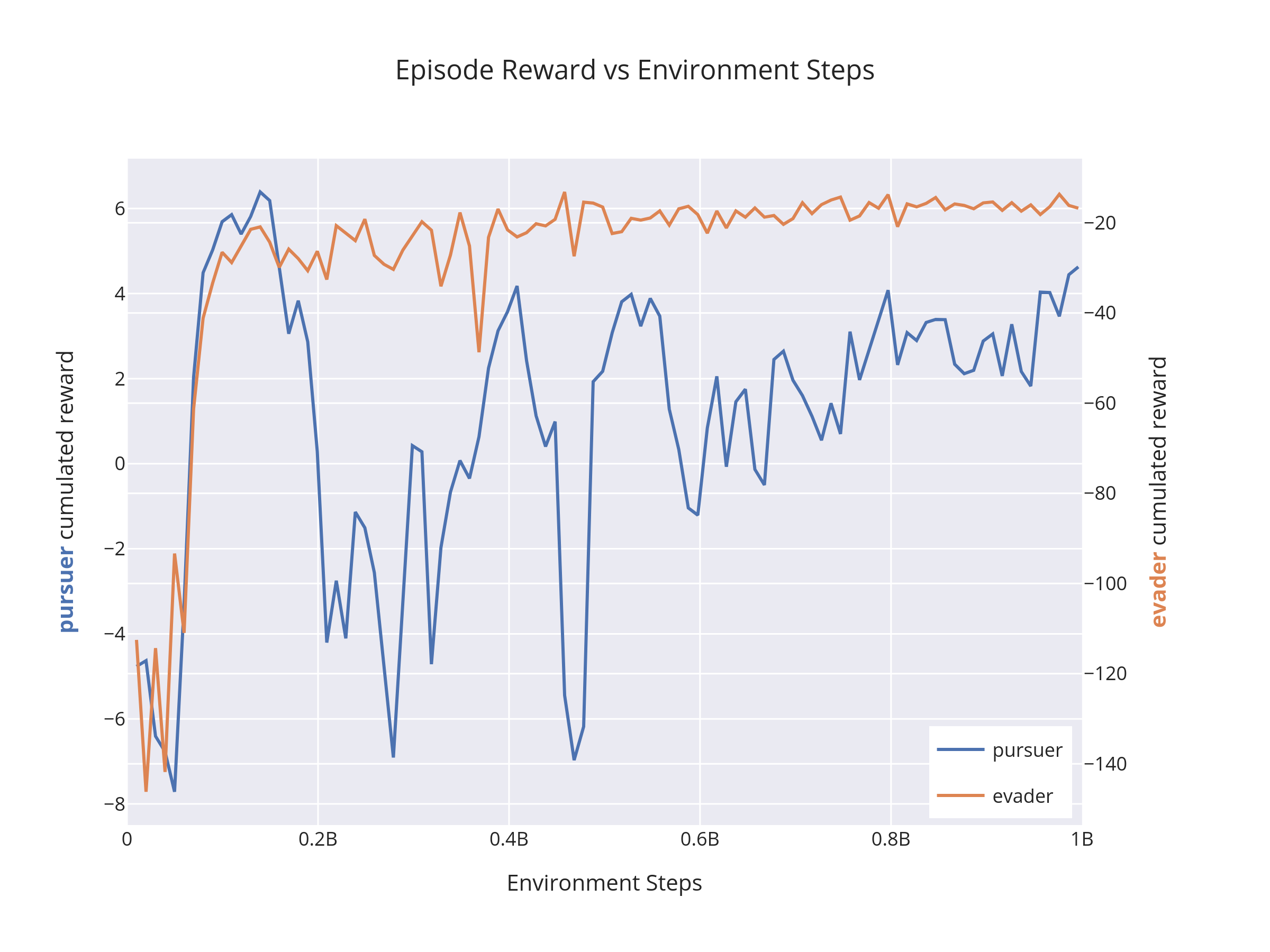} 
    \caption{Average cumulated reward over training.}
    \label{fig:reward}
  \end{subfigure}
  \hfill
  \begin{subfigure}[t]{0.48\textwidth}
    \centering
    \includegraphics[width=\textwidth, clip, trim=0 100 0 200]{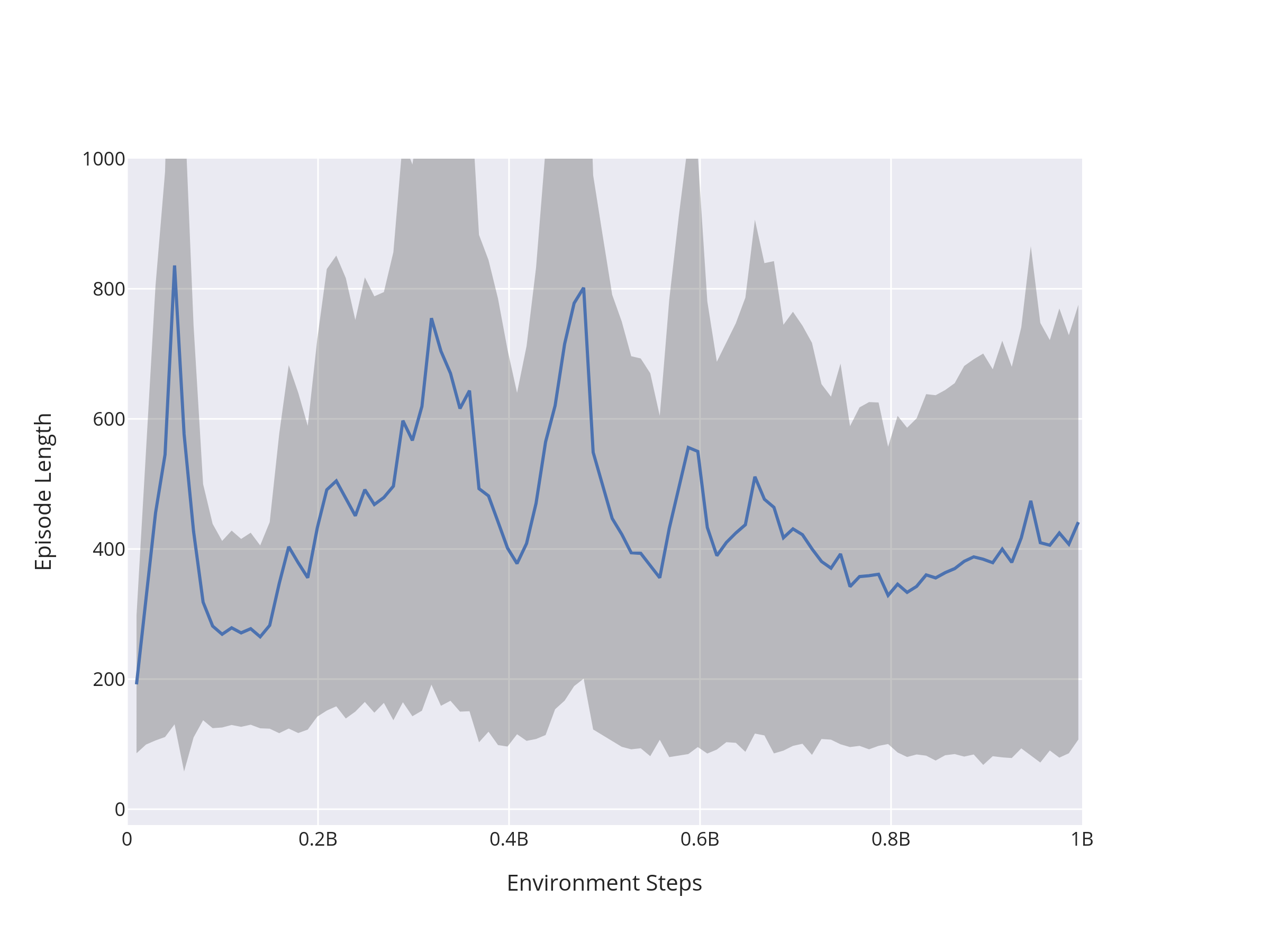}
    \caption{Average episode length.}
    \label{fig:length}
  \end{subfigure}
  \caption{Comparison of learning curves and average episode length.}
  \label{fig:learning_curves}
\end{figure}

Figure~\ref{fig:learning_curves} compares the learning curves of the pursuer and the evader.
The pursuer cumulated return initially rises as the drone learns to fly and avoid crashes and the first interception happens.
Because the evader’s reward contains an additional boundary term, its learning progress is intrinsically slower; it does not reach high-speed flight as early as the pursuer. The pursuer therefore overfits to an increasingly predictable evader.
However, as the evader also learns to fly and evade, the pursuer's return decreases drastically as it can no longer catch the evader.
This also translates into the average episode length which first increases as both agents learn to hover and avoid crashes, but soon falls sharply as the pursuer discovers a quick capture strategy.
Eventually, the pursuer finds a new strategy to catch the evader again, the average episode length decreases and the return of the pursuer increases as it learns to catch the evader more consistently.
This behaviour is typical of co-evolutionary learning \cite{Bansal2017}, and happens multiple times during the training as both agent cycle through periods of adaptation and counter-adaptation.
Both curves converge to a near-stationary value, suggesting that the joint policy profile is approaching a Nash equilibrium.

\subsection{Evaluation in Simulation}

\begin{table}[t]
  \centering
  \caption{Performances of the pursuer and the evader in a 40x40x14m (Large) and a 8x8x5m (Small) arena.}
  \label{tab:performances}
  \begin{tabular}{@{}rc@{\hskip 3pt}c@{\hskip 3pt}cc@{\hskip 3pt}c@{\hskip 3pt}c@{}} \toprule
    \multirow[b]{2}{*}{\textbf{Pursuer mode}}
                             & \multicolumn{6}{c}{\textbf{Evader Mode}}                    \\
    \cmidrule(lr){2-7}
                              & \multicolumn{3}{c}{\textbf{Small arena}}
                             & \multicolumn{3}{c}{\textbf{Large arena}}\\
    
    \midrule
    \multicolumn{1}{c}{\textbf{PP}} & \textbf{Hov.}
                             & \textbf{APF}
                             & \textbf{DRL} & \textbf{Hov.}
                             & \textbf{APF}
                             & \textbf{DRL}              \\
    \cmidrule(lr){1-1}
    	\textbf{Catch Rate (\%)}    & 96.4   & 19.5  & 58.3     & \pursuertextcolor{\textbf{100}}  & 15.9 & 24.6                              \\
    	\textbf{Evade Rate (\%)}    & 0.0    & 47.9  & \evadertextcolor{\textbf{40.0}}     & 0.0                              & 66.6 & \evadertextcolor{\textbf{74.6}}   \\
    of which timeout            & 0.0    & 0.0   & \evadertextcolor{\textbf{0.8}}      & 0.0                              & 0.0  & \evadertextcolor{\textbf{1.1}}                               \\
    \textbf{Crash rates (\%)}                                                                                                             \\
    Pursuer                     & 0.0    & 47.9  & 39.3     & \pursuertextcolor{\textbf{0.0}}  & 53.2 & 73.5                              \\
    Evader                      & -      & 32.0  & \evadertextcolor{\textbf{0.6}}      & -                                & 25.9 & \evadertextcolor{\textbf{0.3}}    \\
    Double                      & 3.6    & 0.6   & 1.0      & \pursuertextcolor{\textbf{0.0}}                              & 0.3  & 0.5                               \\
    \textbf{Time to Catch (s)}                                                                                                            \\
    Mean                        & 2.05   & 8.29  & 5.29     & \pursuertextcolor{\textbf{6.65}} & 8.72 & 8.32                              \\
    Std                         & 1.66   & 3.48  & 4.05     & 3.90                             & 2.94 & 3.11                              \\
    \midrule
    \multicolumn{1}{c}{\textbf{FRPN \cite{pliska2024towards}}} & \textbf{Hov.}
                             & \textbf{APF}
                             & \textbf{DRL} & \textbf{Hov.}
                             & \textbf{APF}
                             & \textbf{DRL}              \\
    \cmidrule(lr){1-1}
  	\textbf{Catch Rate (\%)}   & \pursuertextcolor{\textbf{97.4}}   & 19.6     & 37.7       & 97.5                            & \pursuertextcolor{\textbf{68.8}}  & 49.2                            \\
  	\textbf{Evade Rate (\%)}   & 0.1    & 43.1     & \evadertextcolor{\textbf{59.9}}       & 0.0                             & 1.0                               & \evadertextcolor{\textbf{47.3}}  \\
  of which timeout           & 0.0    & 1.2      & \evadertextcolor{\textbf{0.1}}        & 0.0                             & 0.0                               & \evadertextcolor{\textbf{20.2}}  \\
    \textbf{Crash rates (\%)}                                                                                                                                            \\     
  Pursuer                    & \pursuertextcolor{\textbf{0.1}}    & 42.0     & 59.8       & 0.0 & \pursuertextcolor{\textbf{1.0}}                               & 27.1                             \\
  Evader                     & -      & 36.4     & \evadertextcolor{\textbf{1.3}}        & -                               & 30.2                              & \evadertextcolor{\textbf{3.2}}   \\
  Double                     & \pursuertextcolor{\textbf{2.5}}    & 0.8      & 1.1        & 2.5                             & \pursuertextcolor{\textbf{0.0}}                               & 0.3                              \\
    \textbf{Time to Catch (s)}\\                                                                               
  Mean                       & \pursuertextcolor{\textbf{2.03}}   & 8.49     & 6.88       & \pursuertextcolor{\textbf{2.70}}                            & \pursuertextcolor{\textbf{4.97}}  & 6.72                             \\
  Std                        & 1.33   & 3.15     & 4.03       & 1.29                            & 3.44                              & 3.67                             \\
    \midrule
    \multicolumn{1}{c}{\textbf{DRL (Ours)}}                 & \textbf{Hov.}
                             & \textbf{APF}
                             & \textbf{DRL} & \textbf{Hov.}
                             & \textbf{APF}
                             & \textbf{DRL}              \\
    \cmidrule(lr){1-1}
  	\textbf{Catch Rate (\%)}    & 90.7  & \pursuertextcolor{\textbf{71.8}}   & \pursuertextcolor{\textbf{78.8}}     & 20.7                            & 34.0   & \pursuertextcolor{\textbf{66.5}}          \\
  	\textbf{Evade Rate (\%)}    & 6.6   & 6.9    & \evadertextcolor{\textbf{16.5}}     & \evadertextcolor{\textbf{75.6}} & 50.4  & 31.9                                      \\
  of which timeout            & 1.0   & 0.3    & \evadertextcolor{\textbf{2.3}}      & \evadertextcolor{\textbf{74.0}} & 8.1  & 14.7                                      \\
    \textbf{Crash rates (\%)}                                                                                                                     \\
  Pursuer                     & 5.6   & \pursuertextcolor{\textbf{6.6}}    & \pursuertextcolor{\textbf{14.2}}     & 1.6                             & 42.3   & \pursuertextcolor{\textbf{17.2}}          \\
  Evader                      & -     & 20.3   & \evadertextcolor{\textbf{4.1}}      & -                               & 15.4  & \evadertextcolor{\textbf{1.5}}            \\
  Double                      & 0.4   & \pursuertextcolor{\textbf{1.0}}    & \pursuertextcolor{\textbf{0.6}}      & 3.7                             & 0.2   & \pursuertextcolor{\textbf{0.1}}                                       \\
    \textbf{Time to Catch (s)}                                                                                                                \\ 
  Mean                        & 2.62  & \pursuertextcolor{\textbf{4.18}}   & \pursuertextcolor{\textbf{3.78}}     & 8.96                            & 7.61  & \pursuertextcolor{\textbf{6.62}}          \\
  Std                         & 2.76  & 3.89   & 3.60     & 2.46                            & 3.55  & 3.34                                      \\
    \bottomrule
    \multicolumn{7}{l}{Hov.: Hovering, APF: Artificial Potential Field} \\
    \multicolumn{7}{l}{\pursuertextcolor{\textbf{text in blue}} : best pursuer against this column's evader} \\
    \multicolumn{7}{l}{\evadertextcolor{\textbf{text in orange}} : best evader against this row's pursuer} 
  \end{tabular}
\end{table}

\begin{figure*}[t]
  \centering
  \begin{minipage}[b]{0.56\textwidth} 
    \centering
    \includegraphics[width=\textwidth]{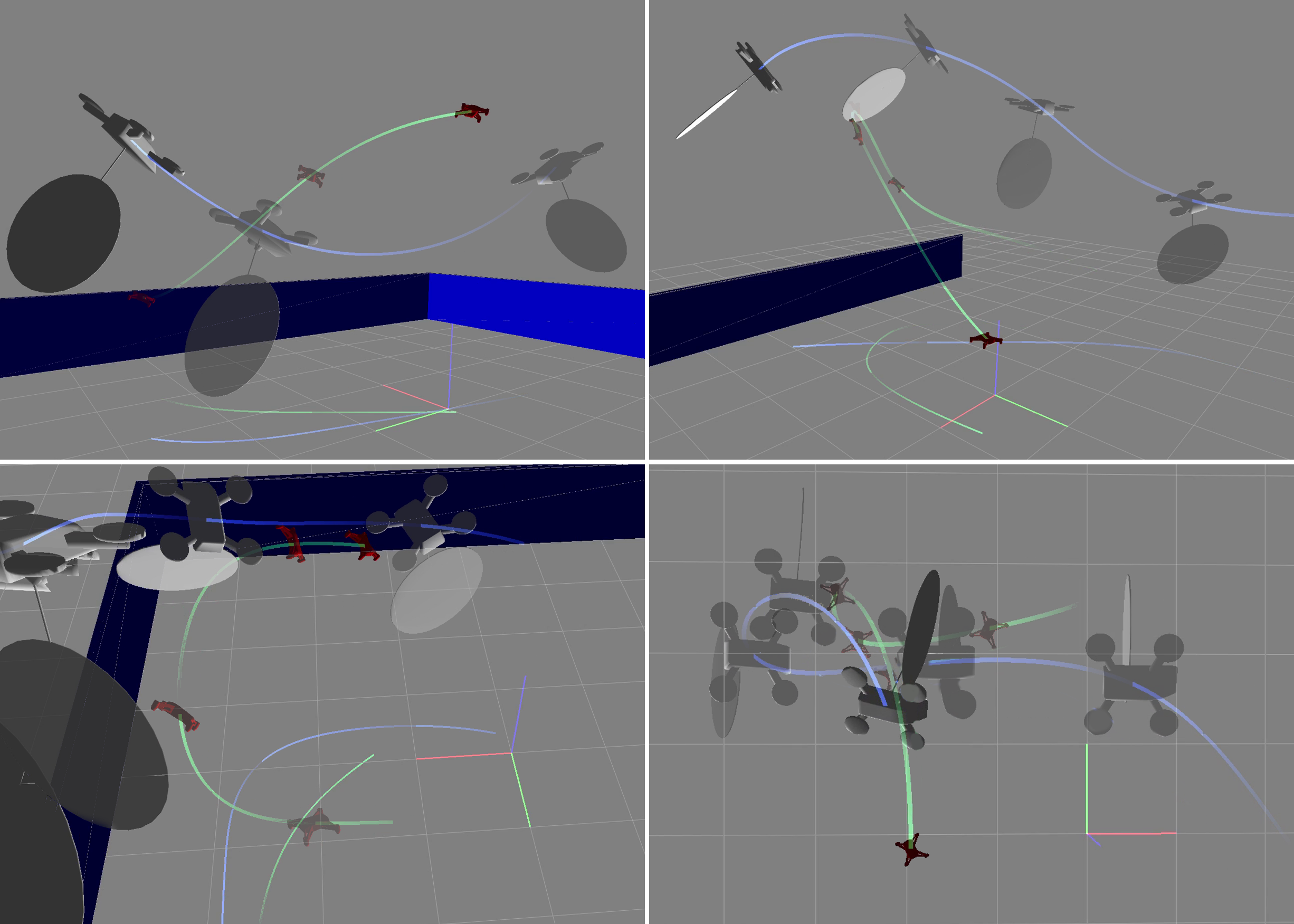}
    \captionsetup{width=0.95\linewidth}
    \caption{Evasive manoeuvres: from top left to bottom right, the evader (green) performs a vertical escape, a dive, a sharp turn, and a sudden stop followed by a feint.}
    \label{fig:qualitative_evader}
  \end{minipage}
  \begin{minipage}[b]{0.40\textwidth} 
    \centering
    \includegraphics[width=\textwidth]{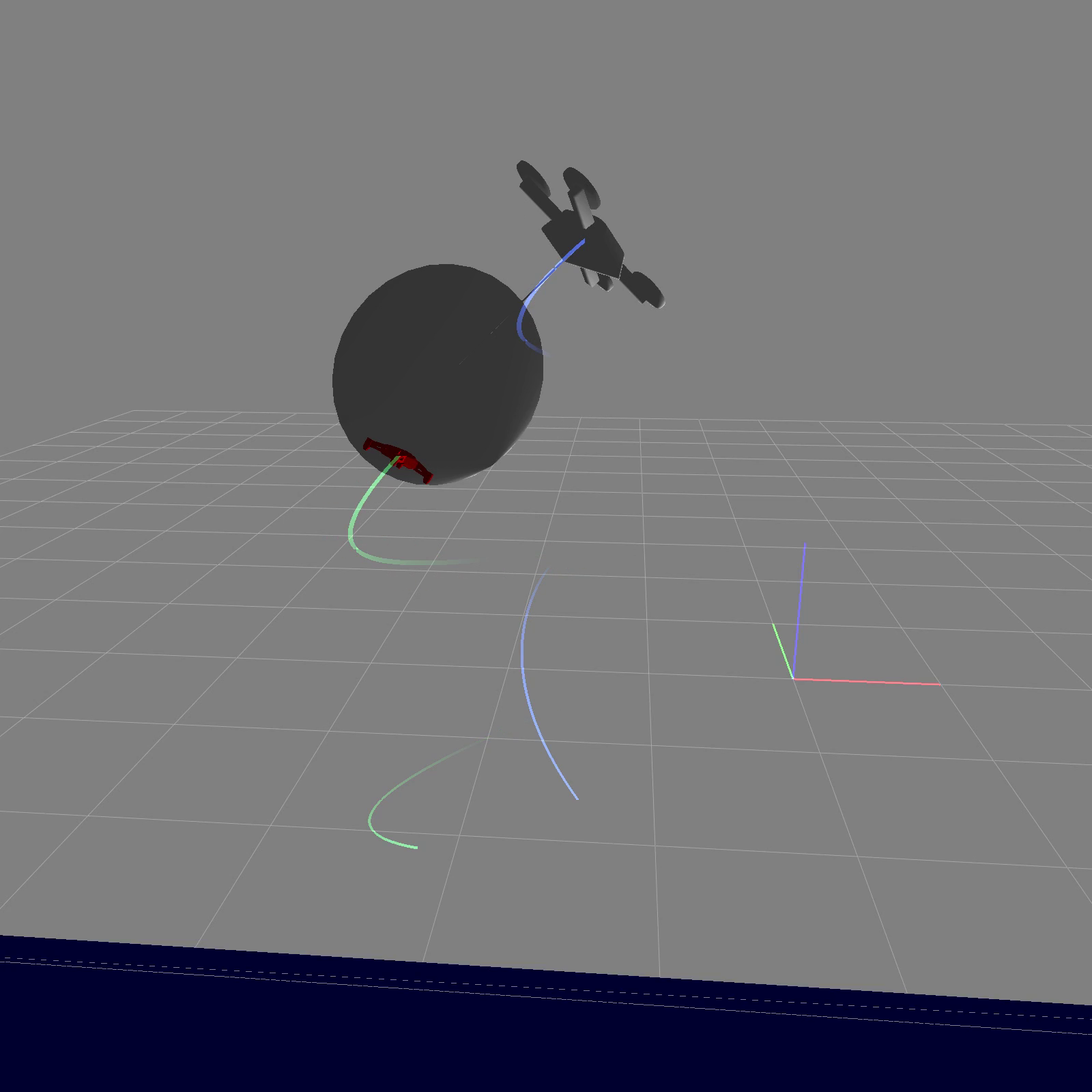}
    \captionsetup{width=0.95\linewidth} 
    \caption{a high roll angle catch: the pursuer (blue) intercepts the evader (green) with a roll angle of more than 45 degrees.}
    \captionsetup{width=\linewidth} 
    \label{fig:qualitative_pursuer}
  \end{minipage}
\end{figure*}

We compare the performances of the trained policies with baseline heuristic methods. 
For the pursuer, Pure-Pursuit (PP), a classical interception strategy where the pursuer follows a straight line towards the position of the evader, and Fast-Response Proportional Navigation \cite{pliska2024towards} which is an evolution of Proportional Navigation for manoeuvring multi-rotors. 
For the evader, a hovering strategy where the evader tries to maintain a fixed position in space, and an Artificial Potential Field strategy where the evader is repelled by the pursuer and the boundaries of the arena. We use the potential field formulation from \cite{zhang2023dacoopadecentralizedadaptivecooperative}.
In the sake of comparison, these heuristic methods only access the position and velocity of the opponent agent, as our RL policies only access this information.
How to estimate the state of the evader in high-speed manoeuvring flights is not in the scope of this paper.
The heuristic baselines give velocity or acceleration commands that are then converted to body rates and collective thrust using the SE(3) controller described in Section~\ref{sec:simulator}.

The main comparison metrics are the catch rate of the pursuer, the evade rate of the evader, the time to catch and the crash rate. 
The catch rate is the percentage of episodes where the pursuer successfully catches the evader before a timeout of 10 seconds. 
The evade rate is the percentage of episodes in which the evader avoids capture for 10 seconds or the pursuer crashes.
We also identify three different crash rates: pursuer crash rate and evader crash rate are the percentage of episodes where either the pursuer or the evader crashes alone, and double-crash rate is the percentage of episodes where both agents crash simultaneously.
Finally, the time to catch is the time taken by the pursuer to catch the evader.

Time-to-catch is naturally biased towards lower values as it only consider successful catches, thus a weaker pursuer can appear to have a better time to catch as it would only succeed in catching the easiest targets without crashing.
To alleviate this issue, we use a right-censored metric for the time to catch: if the episode ends because of a crash or a timeout, the time to catch is considered to be of 10 seconds. 

\begin{figure*}[t]
  \centering
  \includegraphics[width=0.9\textwidth, clip, trim=0 0 0 10]{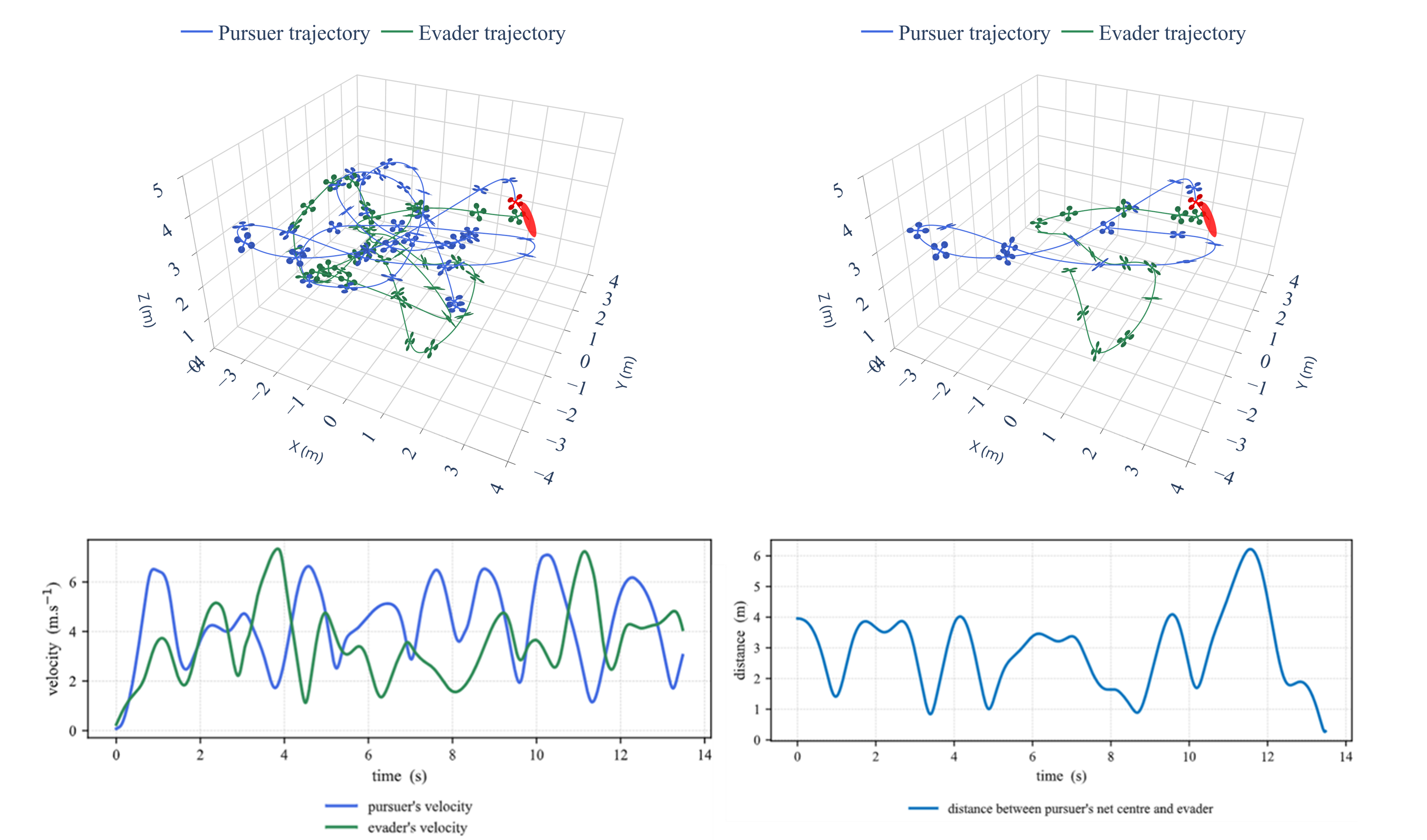}
  \caption{Simulation results: A full pursuit-evasion episode using the trained policies. The trajectory on the right is a shortened version of the full episode for more visibility. We can see the pursuer first missing the evader in a first attempt, then successfully catching it after a second attempt.}
  \label{fig:sim_trajectory}
\end{figure*}

We evaluate the performances of our strategies in two different settings. 
First in a large arena of size $40\times40\times14$ meters, with the evader constrained in a smaller volume of size $20\times20\times4$ meters in the centre of the arena.
In this setting, the agents can reach higher speeds and perform long-range manoeuvres with low risk of crashing into the boundaries.
This is to the advantage of the heuristic pursuer baselines, which do not account for the presences of boundaries.
Then in a smaller arena of size $8\times8\times5$ meters, with the evader constrained in a volume of size $6\times6\times4$ meters in the centre of the arena, closer to indoor voliere flight conditions.
In this setting, the agents are more constrained by the boundaries and have to perform tighter manoeuvres.
For each setting, a specific pursuer and an evader model was trained for this specific arena size.  
For each combination of pursuer and evader strategies, we run 10,000 episodes and report the averaged metrics in Table~\ref{tab:performances}.

The learned evader outperforms the \textit{moving} heuristic evaders in all settings, achieving a higher evade rate and lower crash rate against all pursuers.
The learned evader is particularly effective against the heuristic pursuers, which have a high crash rate when facing agile manoeuvres. 
Against FRPN in the larger arena where it is less crash-prone, half of the successful evasions are due to timeouts, showing that the learned evader can consistently avoid capture for the full duration of the episode.
This shows that we successfully trained an agile evader that can exploit the full 3D space to avoid capture while avoiding crashes.

All the pursuer heuristic baselines performances drop when facing the agile learned evaders. 
Their crash rate is high, as it does not take into account for the presence of boundaries. This effect is exacerbated in the smaller arena. 
In comparison, the learned pursuer that was trained to avoid crashes shows a much lower crash rate against the agile learned evader.
The learned pursuer has also the highest catch rate and lowest time to catch against the agile learned evader, showing that it has learned effective interception strategies against agile manoeuvres while respecting arena boundaries and minimizing crashes.
The low time-to-catch shows that the learned pursuer succeeds against the hardest-to-catch, most time-consuming evaders where the other methods fail or crash.

However, the performances of the learned pursuers drop significantly when facing the heuristic evaders, especially the hovering one.
While the learned pursuer still display a low crash rate, its catch rate is much lower than the heuristic pursuers in the larger arena.
This suggests that the learned pursuer has overfitted to the strategies of the agile learned evader encountered during training, and fails to find effective interception strategies against less agile evaders. 
This is especially true against the hovering evader. It seems to be a very easy target to catch, but this situation was likely not encountered during training, as the evader was always trying to escape.
Moreover, the pursuer's observation do not encode history information, and cannot infer that the evader is stationary from its current observed state (position, velocity). As a result, it did not learn to exploit the lack of movement to optimize its interception strategy. In fact, it must expect it to flee at any moment.
This is less pronounced in the smaller arena, where the boundaries further constrain the evader's movements. 
It is likely that the learned pursuer encountered more trajectories where the evader was close to stationary during training, allowing it to learn some interception strategies against this type of target. 
As a result, the learned pursuer still achieves a higher catch rate and faster time to catch than the heuristic pursuers against all moving evaders in the smaller arena, primarily due to its ability to avoid boundaries and crashes.

\subsection{Qualitative Results in Simulation}

In this section, we present qualitative results of our trained policies in the small arena setting ($8\times8\times5$ m). 

As shown in Figure~\ref{fig:qualitative_evader}, the evader learned a diverse set of agile evasive manoeuvres, including high accelerations, high velocity flights, sudden stops, sharp turns, vertical movements, and feints.
In response, the pursuer learned to anticipate these manoeuvres. We observed the pursuer catching the evader with very high roll and pitch angles ($>$45 degrees) (Figure~\ref{fig:qualitative_pursuer}).
Despite being not specifically enforced by the reward function to turn the heading towards the evader, the pursuer learned to do so in order to maximize the surface of the catching net facing the evader, which increases the chances of a successful capture.

It also learned to catch the evader using both sides of the catching net, to intercept an evader that went behind it without turning around.

One trajectory obtained is shown in Figure~\ref{fig:sim_trajectory}. Both the pursuer and the evader display agile manoeuvres in a very restricted arena, with velocities of up to $\sim7.5$ m/s. The pursuer (in blue) is able to catch the evader (in green) after 13 seconds of intense chase, showcasing the ability of the learned policies to sustain high intensity flights while avoiding crashes.

\subsection{Real-World Demonstration}

We demonstrated the trained policies in a real-world scenario in our indoor flight arena of size $8\times8\times5$ meters.
The policies have been directly transferred from simulation to reality without any additional fine-tuning or adaptation.
For this flight, the evader was simulated on a ground station computer, while the pursuer was flying a real quadcopter equipped with a Betaflight \cite{betaflight2022} flight controller.
The flight logs were recorded, including and action commands, and analysed afterward to identify successful catches and the collisions between the pursuer and the evader.
The state of the real drone is estimated using a motion capture system (OptiTrack) that provides accurate position and orientation data at 200 Hz and transferred to the simulation.
The neural network policy was executed remotely on the ground station computer and the outputted control commands transferred to the drones via an RF link at 100 Hz.
We adopted this Hardware-In-The-Loop setup to ensure safety during the flights, but it is not a limitation of our approach as the trained policies can be executed on-board in a decentralized way.

\begin{figure}[t]
  \centering
  \includegraphics[width=\linewidth]{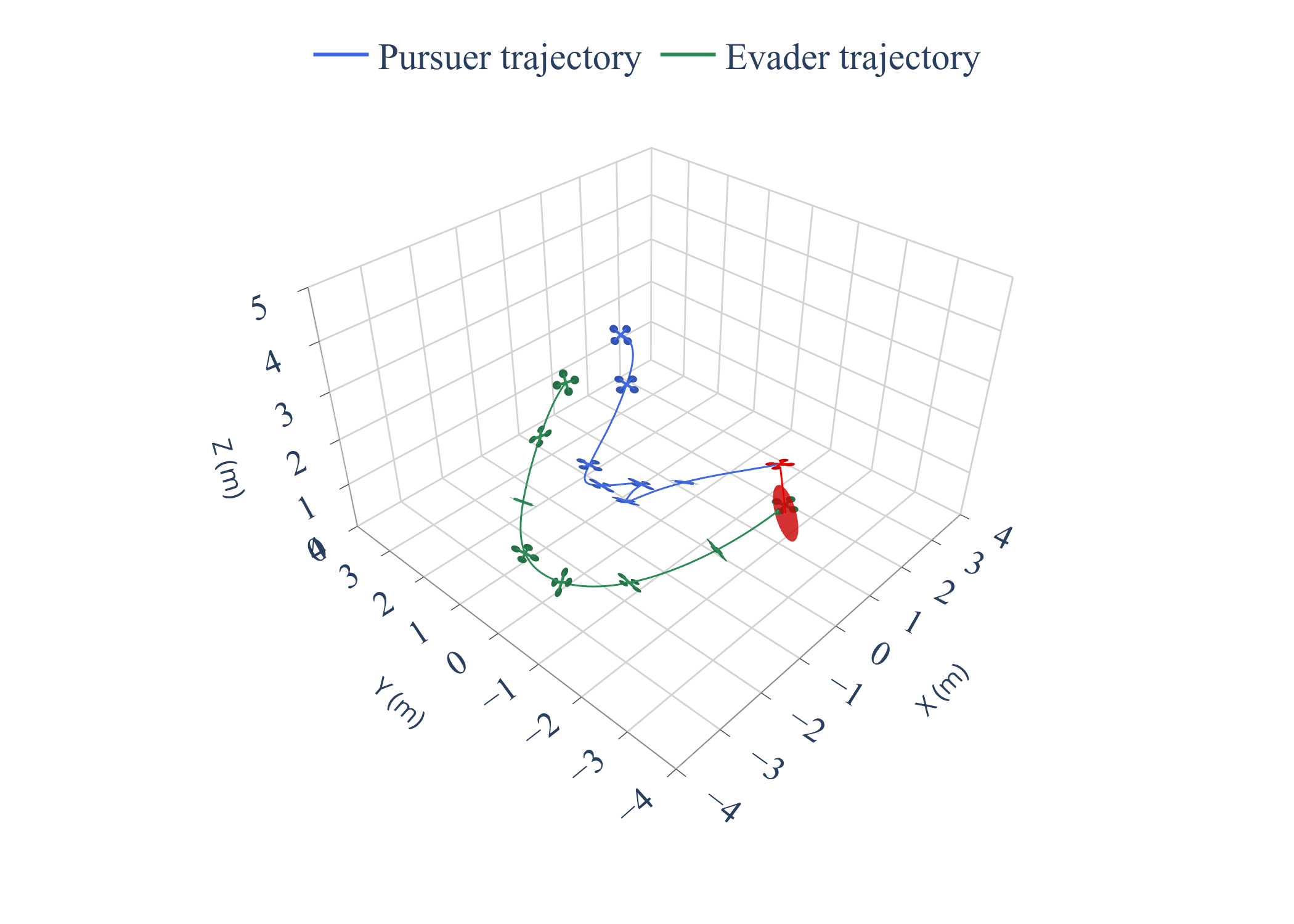}
  \caption{A portion of the real-world flight trajectory. The pursuer (blue) successfully caught the evader (green).}
  \label{fig:real_trajectory}
\end{figure}

The pursuer managed to fly without crashing or exiting the arena during 28 seconds, and successfully caught the evader 7 times during this period. A portion of the recorded trajectory is shown in Figure~\ref{fig:real_trajectory}.

The flight logs were analysed to identify the error between the simulation and the real-world execution. 
At each time step, we computed the error between the expected next state from the simulation and the actual next state recorded from the real-world flight with the actions given to the policy network.
With a period between time steps at 100Hz of 0.01 seconds, the position RMSE is 0.009 m on the xy plane and 0.003 m on the z axis, while the velocity RMSE is 0.070 m/s on the xy plane and 0.040 m/s on the z axis.

\section{Conclusion}
\label{sec:conclusion}

In this work, we addressed the challenging problem of intercepting an agile aerial target using a pursuer drone equipped with a catching net.
We formulated this task as a competitive multi-agent reinforcement learning problem, training independent policies for both the pursuer and the evader using PPO with low-level control inputs (collective thrust and body rates).
A key element of our approach was the integration of a high-fidelity quadrotor dynamics model and a multi-agent reinforcement learning framework for the training of both the pursuer and the evader.

Our simulation results demonstrated that the trained policies outperformed classical heuristic baselines in simulated interception tasks of agile evader, achieving higher catch rates and demonstrating greater robustness against crashes, particularly when facing agile, learned opponents.
Furthermore, the development of our simulation environment entirely within the JAX framework proved crucial, enabling massively parallelized execution and drastically accelerating the training process, which made extensive RL training computationally feasible.

While comprehensive quantitative evaluation in the physical world remains challenging, we successfully demonstrated the learned policies on agile quadrotors in our indoor flight arena, validating the potential for zero-shot sim-to-real transfer and showcasing the practical applicability of our approach.

Overall, this research highlights the effectiveness of Multi-Agent Competitive Reinforcement Learning for generating highly agile and reactive control policies for complex robotic interaction tasks like drone interception.

\section{Limitations} \label{sec:limitations}
While our approach demonstrates promising results for agile drone interception, several limitations should be acknowledged.

First, the sim-to-real gap remains a significant challenge.
Although our simulation uses quadrotor dynamics identified from real flight data, trained policies remain sensitive to mismatches between the simulated model and actual hardware. Uncertainties in parameters like mass, inertia, or motor limits can degrade real-world performance. Incorporating domain randomization during training could improve robustness to such discrepancies.

Second, we observed instances where agents appeared to overfit to certain interaction patterns.
When opponents deviated from typical behaviors (e.g., flying erratically or hovering), agents sometimes responded suboptimally or failed (e.g., crashing), rather than robustly pursuing their objectives. 
When training only against the latest version of the opponent, agents are prone to forget how to deal with previously encountered strategies, and it limits the generalization capabilities of the learned policies.
Expanding the diversity of opponent strategies met during training, for example via Self-Play or Population-Based Training \cite{OpenAI2019}, could mitigate overfitting and improve generalization.

Third, we assume perfect state information for both agents during training and execution. Incorporating realistic sensor models and handling partial observability are important for real-world deployment, where robust perception of the target in high-velocity flights and state estimation are required but were not addressed in this study.

Fourth, the current study focuses on a one-vs-one "dog-fight" scenario within a bounded, obstacle-free arena.
Extending the approach to handle multiple pursuers and/or evaders, operate in cluttered environments, or address different objectives like area defense requires further investigation.

Finally, while we demonstrated feasibility in real-world flights, the quantitative evaluation was primarily conducted in simulation.
A more extensive real-world experimental campaign would be necessary to rigorously quantify performance metrics like catch rate and time-to-catch under physical conditions.

\bibliographystyle{IEEEtran}
\bibliography{shield}

\end{document}